\pdfoutput=1
\documentclass[11pt]{article}
\usepackage{graphicx}
\usepackage{booktabs}
\usepackage{afterpage}
\usepackage{adjustbox}
\usepackage{tabularx}
\usepackage{subfig}
\usepackage{xr}
\usepackage{xspace}
\usepackage{tablefootnote}
\usepackage{amsmath}
\usepackage{hyperref}

\newcommand{\tfew}{\textsc{T-Few}\xspace}
\newcommand{\tfewsmall}{\textsc{T-Few} 3B\xspace}
\newcommand{\tfewlarge}{\textsc{T-Few} 11B\xspace}
\newcommand{\adapet}{\textsc{Adapet}\xspace}
\newcommand{\setfit}{\textsc{SetFit}\xspace}
\newcommand{\setfitbase}{\textsc{SetFit$_{\scriptsize\textsc{MPNet}}$}\xspace}

\newcommand{\setfitlarge}{\textsc{SetFit$_{\scriptsize\textsc{RoBERTa}}$}\xspace}
\newcommand{\setfittiny}{\textsc{SetFit$_{\scriptsize\textsc{MiniLM}}$}\xspace}
\newcommand{\robertalarge}{\textsc{RoBERTa$_{\scriptsize\textsc{Large}}$}\xspace}
\newcommand{\finetune}{\textsc{FineTune}\xspace}

\newcommand{\perfect}{\textsc{Perfect}\xspace}

\usepackage{pgfplots}
\usepackage{pgfkeys}
\definecolor{decentgrey}{RGB}{232,232,232}
\pgfplotsset{compat=1.13}
\definecolor{c0}{cmyk}{1,0.3968,0,0.2588} 
\definecolor{c1}{cmyk}{0,0.6175,0.8848,0.1490} 
\definecolor{c2}{cmyk}{0.1127,0.6690,0,0.4431} 
\definecolor{darkgrey}{RGB}{149,149,149}
\usetikzlibrary{calc,fit,positioning,arrows,arrows.meta,backgrounds,decorations.pathreplacing}
\usepgfplotslibrary{fillbetween}
\definecolor{bg}{rgb}{0.95,0.95,0.95}

\usepackage[]{acl}
\usepackage{times}
\usepackage{latexsym}

\usepackage[T1]{fontenc}

\usepackage[utf8]{inputenc}

\usepackage[verbose=silent]{microtype}

\title{Efficient Few-Shot Learning Without Prompts}

\author{Lewis Tunstall$^1$, Nils Reimers$^2$, Unso Eun Seo Jo$^1$, Luke Bates$^3$,\\ \textbf{Daniel Korat$^4$, Moshe Wasserblat$^4$, Oren Pereg$^4$}  \\
  $^1$Hugging Face \quad $^2$cohere.ai \\
  $^3$Ubiquitous Knowledge Processing Lab, Technical University of Darmstadt \\
  $^4$Emergent AI Lab, Intel Labs \\
  $^1${\tt firstname@huggingface.com} \quad $^2${\tt info@nils-reimers.de}\\
  $^3${\tt bates@ukp.informatik.tu-darmstadt.de} \\
  $^4${\tt firstname.lastname@intel.com} \\
}

\begin{document}
\maketitle

\begin{abstract}
Recent few-shot methods, such as parameter-efficient fine-tuning (PEFT) and pattern exploiting training (PET), have achieved impressive results in label-scarce settings. However, they are difficult to employ since they are subject to high variability from manually crafted prompts, and typically require billion-parameter language models to achieve high accuracy. To address these shortcomings, we propose \setfit (\textbf{Se}ntence \textbf{T}ransformer \textbf{Fi}ne-\textbf{t}uning), an efficient and prompt-free framework for few-shot fine-tuning of Sentence Transformers (ST). \setfit works by first fine-tuning a pretrained ST on a small number of text pairs, in a contrastive Siamese manner. The resulting model is then used to generate rich text embeddings, which are used to train a classification head. This simple framework requires no prompts or verbalizers, and achieves high accuracy with orders of magnitude less parameters than existing techniques. Our experiments show that \setfit obtains comparable results with PEFT and PET techniques, while being an order of magnitude faster to train. We also show that \setfit can be applied in multilingual settings by simply switching the ST body. Our code\footnote{\url{https://github.com/huggingface/setfit}} and datasets\footnote{\url{https://huggingface.co/setfit}} are made publicly available. 
\end{abstract}

\section{Introduction}
Few-shot learning methods have emerged as an attractive solution to label-scarce scenarios, where data annotation can be time-consuming and costly. These methods are designed to work with a small number of labeled training examples, and typically involve adapting pretrained language models (PLMs) for specific downstream tasks. 

Today, there exist several approaches to few-shot learning with PLMs. These include in-context learning (ICL), parameter-efficient fine-tuning (PEFT), and pattern exploiting training (PET). Unfortunately, these approaches can be impractical for many researchers and practitioners. One disadvantage is that these approaches typically rely on the use of large-scale language models to achieve high performance. For example, \tfew \cite{t-few} is based on the 11 billion parameter model T0 \cite{TO_2021}, while GPT-3 \cite{GPT3_2020} is an order of magnitude larger. Secondly, training and deploying these few-shot methods typically requires specialized infrastructure with limited accessibility. 
Moreover, PET and the prominent PEFT methods require, as part of their training, the input of manually generated prompts, yielding varying outcomes depending on the level of manual prompt-engineering.

\begin{figure}[t]%
    \centering
    \includegraphics[width=0.45\textwidth]{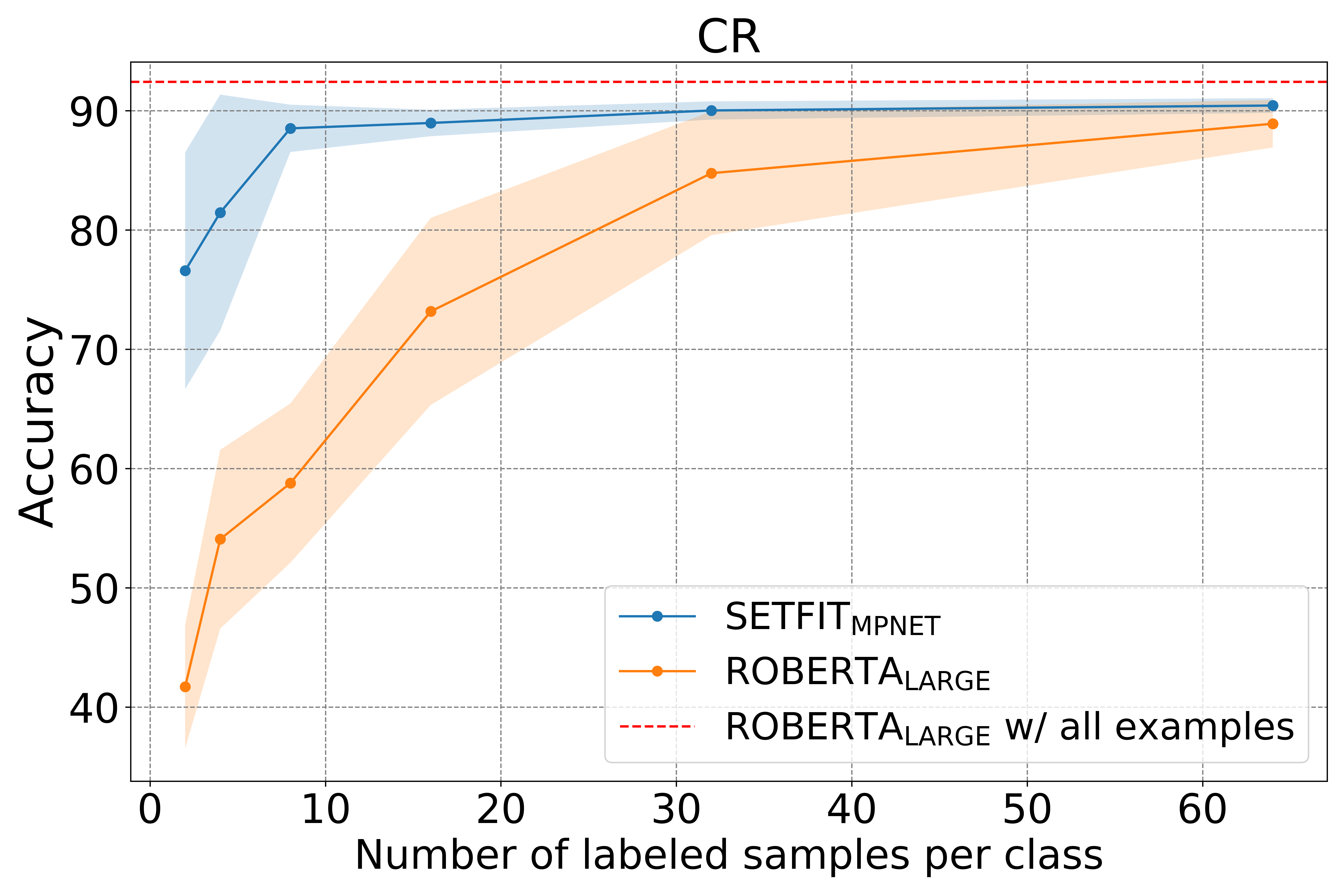}
    \caption{Compared to standard fine-tuning, \setfit is more sample efficient and exhibits less variability when trained on a small number of labeled examples.}
    \label{learning-curves}
\end{figure}

In this paper, we propose \setfit, an approach based on Sentence Transformers (ST) \citep{S-BERT-reimers-gurevych-2019} that dispenses with prompts altogether and does not require large-scale PLMs to achieve high accuracy. For example, with only 8 labeled examples in the Customer Reviews (CR) sentiment dataset, \setfit is competitive with fine-tuning on the full training set, despite the fine-tuned model being three times larger (see Figure~\ref{learning-curves}).

We demonstrate \setfit's efficacy in few-shot text classification over a range of NLP datasets and in multiple scenarios including distillation and non-English data. We compare our method to standard PLM fine-tuning, state-of-the-art PET- and PEFT-based methods such as \adapet \cite{tam-etal-2021-improving} and \tfew \cite{t-few}, as well as recent prompt-free techniques such as \perfect \cite{karimi-mahabadi-etal-2022-prompt}. 

We summarize our contributions as follows:

\begin{enumerate}
    \item We propose \setfit  -- a simple and prompt-free method -- and provide a comprehensive guide for applying it in practical few-shot settings.
    \item We evaluate \setfit's performance on a number of few-shot text classifications tasks 
    and show that it outperforms the state-of-the-art prompt-free method and
    ranks alongside much larger prompt-based, few-shot models.
    \item We make the code and data used in our work publicly available. 
\end{enumerate}

\section{Related Work}
\setfit engages with two related lines of literature. We first extend the small but significant body of work on Sentence Transformers (ST) for text classification. \citet{Sent_emb_2018} introduced the idea of using sentence embeddings for text classification.
\citet{Sent_tranformers_2021} used 'out-of-the-box' STs for text classification without fine-tuning them. \setfit differs from these works in two aspects:  First, we fine-tune the ST in a Siamese manner for a text classification objective showing that it significantly enhances performance; second, we demonstrate this approach in few-shot setups.

\setfit is also related to the recently emerging few-shot and zero-shot training line of literature as 
few-shot and zero-shot approaches have recently received a great deal of interest in the research community due to the availability of pretrained language models and the untapped capacity to use them in resource-constrained domains. Specifically, we discuss ICL, PEFT, and prompt-based fine-tuning.

ICL models directly generate predictions based on input-to-output training examples provided as prompts, without any parameter updates.  Perhaps the best known example is GPT-3 \citep{DBLP:journals/corr/abs-2005-14165}, which achieves remarkable few-shot performance. However, GPT-3 contains 175 billion parameters and requires massive computational resources, prompt engineering, and can only utilize pretrained knowledge.

PEFT methods, such as adapters \citep{adapter}, hold the majority of parameters fixed during training and only update small feed-forward networks that are inserted within the larger model architecture. A recent example is \tfew \citep{t-few}, which outperforms GPT-3 at much lower computational cost. It accomplishes this by adding learned vectors that rescale the network's internal activations. \tfew is 16 times smaller than GPT-3, but is still too large to be utilized as a practical tool in industry. It also requires a set of handcrafted prompts for each dataset.

Another alternative to ICL is prompt-based fine-tuning. This approach converts the downstream classification task into a masked-language modeling (MLM) objective. The model outputs tokens in a cloze-style format that maps to the corresponding labels via a predefined template.
A well known example of this method is Pattern Exploiting Training (PET) \citep{schick-schutze-2021-just, schick-schutze-2021-exploiting} . Like GPT-3, PET relies on manually-crafted prompts, but since the model can be fine-tuned to specific tasks, PET-based approaches typically outperform GPT-3 in few-shot scenarios, even with far smaller PLM backbones. PET has since been extended in two main directions: \adapet \citep{tam-etal-2021-improving}, which improves PET with a decoupled label objective and label-conditioned MLM objective, and \perfect \citep{PERFECT} which uses task-specific adapters \citep{houlsby2019,pfeiffer-etal-2021-adapterfusion} and multi-token label-embeddings eliminate task prompts and verbalizers.

\begin{figure*}[t]
    \centering
    \includegraphics[scale=0.45]{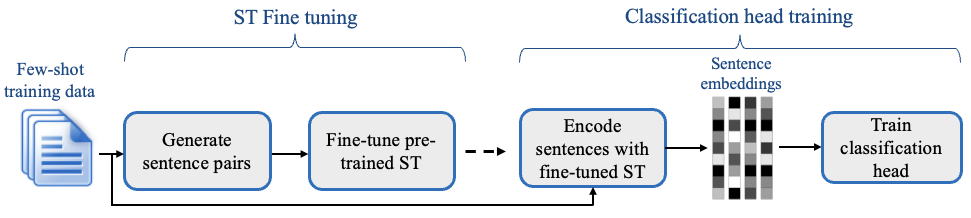}
    \captionof{figure}{\setfit’s fine-tuning and training block diagram.}
    \label{SetFit_fig}
\end{figure*}

\section{SetFit: Sentence Transformer Fine-Tuning}
\setfit is based on Sentence Transformers \cite{S-BERT-reimers-gurevych-2019} which are modifications of pretrained transformer models that use Siamese and triplet network structures to derive semantically meaningful sentence embeddings. The goal of these models is to minimize the distance between pairs of semantically similar sentences and maximize the distance between sentence pairs that are semantically distant. Standard STs output a fixed, dense vector that is meant to represent textual data and can then be used by machine learning algorithms.

\subsection{The \setfit approach for few-shot text classification}

\setfit uses a two-step training approach in which we first fine-tune an ST and then train a classifier head. In the first step, an ST is fine-tuned on the input data in a contrastive, Siamese manner on sentence pairs. In the second step, a text classification head is trained using the encoded training data generated by the fine-tuned ST from the first step.
Figure~\ref{SetFit_fig} illustrates this process, and we discuss these two steps in the following sections.

\paragraph{ST fine-tuning}
To better handle the limited amount of labeled training data in few-shot scenarios, we adopt a contrastive training approach that is often used for image similarity \cite{koch2015siamese}. Formally, given a small set of $K$ labeled examples $D=\{(x_i, y_i)\}$, where $x_i$ and $y_i$ are sentences and their class labels, respectively. 
For each class label $c \in C$, we generate a set of $R$ positive triplets $T^c_p=\{(x_i, x_j, 1)\}$, where $x_i$ and $x_j$ are pairs of randomly chosen sentences from the same class $c$ such that $(y_i=y_j=c)$. Similarly, we also generate a set of $R$ negative triplets $T^c_n=\{(x_i, x_j, 0)\}$, where $x_i$ are sentences from class $c$
and $x_j$ are randomly chosen sentences from different classes such that $(y_i=c,~y_j\neq c)$. Finally, the contrastive fine tuning data set $T$ is produced by concatenating the positive and negative triplets across all class labels; $T=\{(T^0_p, T^0_n), (T^1_p, T^1_n),...,(T^{|C|}_p, T^{|C|}_n)\}$, where $|C|$ is the number of class labels, $|T|=2R|C|$ is the number of pairs in $T$ and $R$ is a hyperparameter. Unless stated otherwise, we used $R=20$ in all the evaluations.

This contrastive fine-tuning approach enlarges the size of training data in few-shot scenarios. Assuming that a small number ($K$) of labeled examples are given for a binary classification task, the potential size of the ST fine-tuning set $T$ is derived from the number of unique sentence pairs that can be generated, namely $K(K-1)/2$, which is significantly larger than just $K$.

\paragraph{Classification head training}
In this second step, the fine-tuned ST encodes the original labeled training data $\{x_i\}$, yielding a single sentence embedding per training sample; $Emb^{x_i}=ST(x_i)$ where $ST()$ is the function representing the fine-tuned ST.
The embeddings, along with their class labels, constitute the training set for the classification head $T^{CH}=\{(Emb^{x_i}, y_i)\}$ where $|T^{CH}|=|D|$. A logistic regression model is used as the text classification head throughout this work. 

\paragraph{Inference}
At inference time, the fine-tuned ST encodes an unseen input sentence $(x_i)$ and produces a sentence embedding. Next, the classification head that was trained in the training step, produces the class prediction of the input sentence based on its sentence embedding. Formally this is $x^{pred}_i=CH(ST(x_i))$, where $CH$ represents the classification head prediction function. 

\section{Experiments}
\label{sec:experiments}

\subsection{Data}
\label{sec:data_section}
We conduct experiments on available text classification datasets. We split the datasets into development and test datasets (See Table \ref{table:english_dev_data} in Appendix \ref{sec:appendix}). The development datasets are utilized for setting \setfit's hyperparameters such as the number of training pairs ($|T|$), the loss function and the optimal number of training epochs.
In order to test the robustness of \setfit to various types of text, we choose test datasets that represent different text classification tasks with a varying number of classes. 
All datasets used are available on the Hugging Face Hub under the \setfit organisation.\footnote{\href{https://huggingface.co/SetFit}{huggingface.co/SetFit}}
In addition we evaluate \setfit on the RAFT benchmark \cite{alex2021raft}, a real-world few-shot text-classification benchmark composed of 11 practical tasks, where each task has only 50 training examples.

\subsection{\setfit models}
We evaluate three variations of \setfit each one uses different underlying model of different size (Shown in Table~\ref{table:setfit_underlying_models})

\begin{table}[ht]
    \begin{minipage}{\linewidth}
	\centering
	\small
    \begin{tabular}{p{2.1cm}ccccc} 
    \toprule[\heavyrulewidth]
    \textbf{Variation} & \textbf{Underlying ST Model} & \textbf{Size$^*$}   \\ 
    \midrule
    \setfitlarge & all-roberta-large-v1\tablefootnote{\url{https://huggingface.co/sentence-transformers/all-roberta-large-v1}} & 355M     \\
    \setfitbase & \small{paraphrase-mpnet-base-v2}\tablefootnote{\url{https://huggingface.co/sentence-transformers/paraphrase-mpnet-base-v2}} & 110M  \\
    \setfittiny & paraphrase-MiniLM-L3-v2\tablefootnote{\url{https://huggingface.co/sentence-transformers/paraphrase-MiniLM-L3-v2}} & 15M   \\

    \bottomrule[\heavyrulewidth] 
    \end{tabular}
	\caption{\setfit model variations using three different underlying ST models. $^*$Number of parameters.}
	
	\label{table:setfit_underlying_models}
	\end{minipage}
\end{table}

\subsection{Baselines}
We compare \setfit's performance against standard transformer fine-tuning and recent best-performing few-shot approaches: \adapet \cite{tam-etal-2021-improving}, \perfect \cite{PERFECT}, and \tfew \cite{t-few}.

\paragraph{Standard fine-tuning}
Our first baseline is \robertalarge \cite{liu2019roberta}, a standard, encoder-only transformer that is fine-tuned for sequence classification. 
Since we assume no validation sets, we constructed validation splits by randomly selecting equally sized portions from the train split. We perform a hyperparameter search on the number of epochs in the range [25,75] and pick the best performing model on a validation split. 

We use a learning rate of $2e^{-5}$ and batch size of 4 in all our experiments. 

\paragraph{\adapet}
Pattern exploiting training (PET) \cite{schick-schutze-2021-just, schick-schutze-2021-exploiting} is a method for improving PLM performance in few-shot setups on downstream tasks by converting textual input into a cloze-style question intended to be reminiscent of the masked language modelling (MLM) objective under which large PLMs such as BERT \cite{devlin-etal-2019-bert} are trained. To determine \setfit's performance relative to PET-based approaches, we compare our method to \adapet \cite{tam-etal-2021-improving}, an extension of PET. In recent work \cite{PET-RAFT-2021}, the authors show that PET-based classification methods excel on the RAFT benchmark, placing second only to much larger models such as \tfew. In our experiments, we used \adapet with default hyperparameters and examined its performance with different PLM backbones, reporting the PLM which resulted in the best performance, albert-xxlarge-v2 \footnote{\url{https://huggingface.co/albert-xxlarge-v2}} (see Appendix \ref{appdx:adapet_training_procedure} in the Appendix for further details). 

\paragraph{\perfect}
\perfect \cite{PERFECT} is another cloze-based fine-tuning method, but unlike PET or \adapet, it does not require handcrafted task prompts and verbalizers. Instead, \perfect uses task-specific adapters \cite{houlsby2019,pfeiffer-etal-2021-adapterfusion} and multi-token label-embeddings which are independent from the language model vocabulary during fine-tuning. To run \perfect on our test datasets, we adapted the configurations provided in the \perfect codebase. 

\paragraph{\tfew}
\tfew \citep{t-few} is a PEFT-based few-shot learning method based on T0 \citep{TO_2021}. The authors provide two versions of \tfew: 11 and 3 billion parameters. Due to compute constraints, we were unable to run the 11 billion version, which requires an 80GB A100 GPU. Running tests on \tfew as opposed to \setfit posed several hurdles. 
First, because \tfew's performance varies significantly depending on the input prompts, we run each experiment using 5 random seeds, and report the median result, as in the original paper. 
Second, \tfew relies on dataset-specific prompts, made available on P3 (Public Pool of Prompts) \citep{p3}. Only one of our test datasets had prompts in P3. For the rest of the datasets, we adapt standardized P3 prompts of similar tasks or implement prompts ourselves (See Appendix \ref{appdx:tfew_prompts}).

\subsection{Experimental Setup}
\label{subsec:main_experimental_setup}
Systematically evaluating few-shot performance can be challenging, because fine-tuning on small datasets may incur instability \citep{dodge2020fine, zhang2021revisiting}. To address this issue, in our experiments we use 10 random training splits for each dataset and sample size. These splits are used as training data across all tested methods. For each method, we report the average measure (depending on the dataset) and the standard deviation across these splits. 
We fine-tune \setfit's ST model using cosine-similarity loss with a learning rate of \(1e^{-3}\), a batch size of 16 and a maximum sequence length of 256 tokens, for 1 epoch.

\section{Results}
\label{sec:main_results}

\begin{table*}[th]
\centering
\begin{tabular}{llllllll}
\toprule
\textbf{Method}             & \textbf{SST-5} &\textbf{AmazonCF} & \textbf{CR} & \textbf{Emotion} & \textbf{EnronSpam} & \textbf{AGNews} & \textbf{Average$^{\dagger}$}\\
\midrule
\multicolumn{7}{c}{\textit{$|N|=8^{*}$}}\\
\finetune\                  & 33.5$_{2.1}$  & 9.2$_{4.9}$   & 58.8$_{6.3}$  & 28.7$_{6.8}$  & 85.0$_{6.0}$  & 81.7$_{3.8}$  & 43.0$_{5.2}$ \\
\perfect\                   & 34.9$_{3.1}$  & 18.1$_{5.3}$  & 81.5$_{8.6}$  & 29.8$_{5.7}$  & 79.3$_{7.4}$          & 80.8$_{5.0}$ & 48.7$_{6.0}$ \\
\adapet\                    & 50.0$_{1.9}$  & 19.4$_{7.3}$  & 91.0$_{1.3}$  & 46.2$_{3.7}$ & 85.1$_{3.7}$  & \textbf{85.1}$_{2.7}$ & 58.3$_{3.6}$ \\
\tfewsmall\                 & \textbf{55.0}$_{1.4}^{\star}$  & 19.0$_{3.9}$  & {\bf 92.1}$_{1.0}$ & {\bf 57.4}$_{1.8}$  & {\bf 93.1}$_{1.6}$  & -- & \textbf{63.4}$_{1.9}$ \\
\setfitbase\                & 43.6$_{3.0}$  & {\bf 40.3}$_{11.8}$ & 88.5$_{1.9}$  & 48.8$_{4.5}$  & 90.1$_{3.4}$  & 82.9$_{2.8}$  & 62.3$_{4.9}$ \\

\midrule
\multicolumn{7}{c}{\textit{$|N|=64^{*}$ }}\\
\finetune\                  & 45.9$_{6.9}$  & 52.8$_{12.1}$ & 88.9$_{1.9}$  & 65.0$_{17.2}$ & 95.9$_{0.8}$  & 88.4$_{0.9}$  & 69.7$_{7.8}$ \\
\perfect\                   & 49.1$_{0.7}$  & {\bf 65.1}$_{5.2}$  & 92.2$_{0.5}$  & 61.7$_{2.7}$  & 95.4$_{1.1}$ & {\bf 89.0}$_{0.3}$ & 72.7$_{1.9}$ \\
\adapet\                    & 54.1$_{0.8}$  & 54.1$_{6.4}$ & 92.6$_{0.7}$  & 72.0$_{2.2}$  & 96.0$_{0.9}$  & 88.0$_{0.6}$ & 73.8$_{2.2}$ \\
\tfewsmall\                 & \textbf{56.0}$_{0.6}$   & 34.7$_{4.5}$  & {\bf 93.1}$_{1.0}$ & 70.9$_{1.1}$  & {\bf 97.0}$_{0.3}$  & --   & 70.3$_{1.5}$ \\
\setfitbase\                    & 51.9$_{0.6}$  & 61.9$_{2.9}$  & 90.4$_{0.6}$  & {\bf 76.2}$_{1.3}$  & 96.1$_{0.8}$  & 88.0$_{0.7}$  & \textbf{75.3}$_{1.3}$ \\
\midrule\midrule
\multicolumn{7}{c}{\textit{$|N|=$ Full$^{**}$}}\\
\finetune\                  & 59.8          & 80.1          & 92.4          & 92.6         & 99.0  & 93.8          & 84.8 \\
\bottomrule
\end{tabular}

\caption{\label{table:main_results} \setfit performance score and standard deviation compared to the baselines across 6 test datasets for three training set sizes $|N|$. $^{*}$Number of training samples per class. $^{**}$Entire available training data used. $^\dagger$The AGNews dataset is excluded from the average score to enable fair comparison with \tfew (which has AGNews in its training set). $^{\star}$The inputs of SST-5 (but not its labels) appeared in \tfew's training set, as part of Rotten Tomatoes dataset.}
\end{table*}

Table \ref{table:main_results} shows a comparison between \setfitbase and the baselines for $N=8$ and $N=64$ labeled training samples per class. For reference purposes, standard fine-tuning results using the full training data are also shown (in all cases higher scores indicates stronger performance; see Table~\ref{table:english_dev_data} in Appendix \ref{sec:appendix} for dataset metric details). We find that \setfitbase significantly outperforms the \finetune baseline for $N=8$ by an average of 19.3 points. However, as the number of training samples increases to $N=64$, the gap decreases to 5.6 points. 

Similarly, we find that \setfitbase outperforms \perfect by 13.6 and 2.6 points. \setfitbase also outperforms \adapet by 4.0 and 1.5 points for $N=8$ and $N=64$ respectively.
For $N=8$, \setfitbase is on par with \tfewsmall whereas for $N=64$ \setfitbase outperforms \tfewsmall by 5 points on average, despite being prompt-free and more than 27 times smaller.

\paragraph{RAFT results}

\begin{table}
    \begin{tabularx}{\columnwidth}{clcc}
        \toprule
            Rank    & Method            & Score  & Size$^*$\\
            \hline
            1       & \textsc{Yiwise}   & 76.8  &  -\\
            2       & \tfew 11B \            & 75.8  & 11B  \\
            4       & Human baseline    & 73.5  &  - \\
            6       & \setfitlarge\          & 71.3  & 355M  \\
            9       & PET\              & 69.6  & 235M  \\
            11      & \setfitbase\      & 66.9  & 110M \\
            12      & GPT-3\            & 62.7  & 175B \\
        \bottomrule
    \end{tabularx}
    
    \caption{\setfit compared to prominent methods on the RAFT leaderboard (as of Sept. 5, 2022). $^*$Number of parameters.}
    \label{table:raft} 
\end{table}

The test datasets listed in Table~\ref{table:main_results} were not specifically designed for few-shot benchmarking. In order to better benchmark \setfit, we used the RAFT benchmark \cite{alex2021raft} which is specifically designed for benchmarking few-shot methods.
Table \ref{table:raft} shows the average accuracy of \setfitbase and \setfitlarge and four prominent methods. \setfitlarge outperforms GPT3 and PET by 8.6 and 1.7 points respectively while alleviating the need for hand crafting prompts. It also surpasses the human baseline in 7 out of 11 tasks. \setfitlarge falls short of \tfewlarge by 4.5 points. however, \setfitlarge is more than 30 times smaller than \tfewlarge, does not require manual prompt crafting and is much more efficient in training and inference (see Table \ref{table:computation_cost}).

\section{Multilingual Experiments}
To determine \setfit's performance in a multilingual, few-shot text classification scenario, we conducted development and test experiments on multilingual datasets and compared \setfit to standard transformer fine-tuning and \adapet. To the best of our knowledge, this is the first work to examine \adapet on non-English data (see Appendix \ref{sec:appendix} for details). 

\paragraph{Experimental Setup}
For the multilingual experiments, we use the Multilingual Amazon Reviews Corpus (MARC) \cite{marc_2020}. This dataset consists of Amazon reviews in six languages (English, Japanese, German, French, Spanish, and Chinese), where each review is labeled according to a 5-star rating scale. We chose this corpus for its typological diversity in order to examine the generalizability of \setfit and other methods across a variety of languages. 

For the \setfit\ underlying model, we use  paraphrase-multilingual-mpnet-base-v2,\footnote{\url{huggingface.co/sentence-transformers/paraphrase-multilingual-mpnet-base-v2}} which is a multilingual version of paraphrase-mpnet-base-v2 that is trained on parallel data in over 50 languages. 

For the \finetune\ and \adapet baselines, we use \textsc{XLM-RoBERTa$_{\small\textsc{base}}$} \citep{xlm-roberta},\footnote{\url{huggingface.co/xlm-roberta-base}} which has a similar size to the \setfit\ model. We compare the performance of each method using the same settings as \citep{xlm-roberta}: 

\begin{itemize}
    \item \textbf{each:} Train and evaluate on monolingual data to measure per-language performance.
    \item \textbf{en:} Train on the English training data and then evaluate on each language's test set.
    \item \textbf{all:} Train on all the training data and evaluate on each language's test set.
\end{itemize}

\paragraph{Method}
For \setfit\, standard fine-tuning, and \adapet, we adopt the same methodology and hyperparameters used for the monolingual English experiments in \ref{sec:experiments}. We evaluate each method in the few-shot regime ($N=8$ samples per class) and compare against performance of fine-tuning on the full training set of 20,000 examples.

\paragraph{Results}
Table \ref{table:multiling_results} shows the results of \setfit\, standard fine-tuning, and \adapet on each language in MARC, where a higher MAE indicates weaker performance. In the few-shot regime of $N=8$ samples per class, we find that \setfit\ significantly outperforms \finetune\ and \adapet in all settings (each, en, all), with the best average performance obtained when training on English data only.

\begin{table*}[t]
\centering
\begin{tabular}{lclllllll}
\toprule
\textbf{Method} & \textbf{Train} & \textbf{En} & \textbf{De} &\textbf{Ja} & \textbf{Zh} & \textbf{Fr} &\textbf{Es} & \textbf{Average} \\
\midrule
\multicolumn{9}{c}{\textit{$|N|=8^{*}$}} \\
 & each & $122.9_{14.0}$ & $119.9_{13.6}$ & $120.5_{8.0}$ & $128.6_{10.7}$ & $123.2_{13.0}$ & $116.3_{8.3}$ & $121.9_{11.3}$ \\
\finetune\ & en & $115.9_{11.3}$ & $115.2_{12.0}$ & $121.6_{12.3}$ & $123.0_{8.8}$ & $117.3_{13.0}$ & $113.1_{12.4}$ & $117.7_{11.6}$ \\
 & all & $117.8_{4.9}$ & $116.3_{9.7}$ & $121.5_{12.4}$ & $120.5_{6.7}$ & $117.3_{9.9}$ & $110.1_{9.5}$ & $117.2_{8.8}$ \\
 \midrule
 & each & 129.9$_{13.6}$ & 136.4$_{10.6}$ & 130.4$_{13.4}$ & 135.0$_{10.9}$ & 141.8$_{10.1}$ & 136.0$_{10.4}$ & 134.9$_{11.5}$ \\
\adapet\ & en & 138.9$_{17.8}$ & 151.5$_{17.8}$ & 160.8$_{16.7}$ & 158.8$_{16.3}$ & 152.0$_{15.7}$ & 149.8$_{17.1}$ & 152.0$_{16.9}$ \\
 & all & 150.8$_{12.0}$ & 136.2$_{7.0}$ & 150.8$_{10.0}$ & 152.8$_{10.2}$ & 140.0$_{14.0}$ & 145.1$_{4.5}$ & 146.0$_{11.3}$ \\
\midrule
 & each & $82.9_{4.3}$ & ${\bf 80.0}_{2.4}$ & $95.5_{2.8}$ & $95.3_{2.8}$ & $85.3_{6.0}$ & ${\bf 80.8}_{5.4}$ & $86.6_{4.9}$ \\
\setfit\ & en & ${\bf 82.6}_{4.8}$ & $83.4_{5.9}$ & ${\bf 93.2}_{6.6}$ & ${\bf 93.9}_{3.6}$ & ${\bf 82.2}_{4.8}$ & $83.4_{5.9}$ & ${\bf 86.4}_{5.2}$ \\
 & all & $83.0_{5.3}$ & $84.0_{7.6}$ & $97.1_{9.2}$ & $97.4_{6.5}$ & $83.5_{6.5}$ & $84.9_{6.1}$ & $88.3_{6.9}$ \\
\midrule\midrule
\multicolumn{9}{c}{\textit{$|N|=$Full$^{**}$}} \\
 & each & $46.2$ & ${\bf 43.7}$ & ${\bf 46.8}$ & ${\bf 56.6}$ & ${\bf 47.8}$ & ${\bf 45.3}$ & ${\bf 47.7}$ \\
\finetune\ & en & ${\bf 46.1}$ & 46.6 & 61.0 & 69.4 & 55.6 & 52.9 & 55.3 \\
 & all & 46.6 & 49.4 & 61.0 & 69.4 & 55.6 & 55.0 & 56.2 \\
\bottomrule
\end{tabular}
\caption{\label{table:multiling_results}
Average performance (MAE $\times$ 100) on the Multilingual Amazon Reviews Corpus for two training set sizes $|N|$. $^{*}$ No. of training samples per class. $^{**}$Entire available training data used (20,000 samples).
}
\end{table*}

\section{\setfit Model Efficiency}

\subsection{Few-shot distillation}

\label{graph:distillation_graph}
\begin{figure*}

    \begin{tikzpicture}[]
		\begin{axis}[
		width=0.37\linewidth, 
		legend style={draw=darkgrey, fill=white!75, text opacity =1, fill opacity=0.8,
		at={(0.97,0.03)},anchor=south east, font=\sffamily\scriptsize},
		legend cell align=left,
		legend columns=1,
		xtick={1,2,3,4,5,6,7,8},
		xticklabels={8,16,32,64,100,200,1K},
		ylabel=Average Accuracy,
		xlabel={\small Unlabeled Training Set Size ($N$)},
		grid=major, clip=false,
		major grid style={line width=.2pt,draw=decentgrey},
		minor tick style={decentgrey!0},
		x tick label style={/pgf/number format/1000 sep=},
		tick label style={font=\tiny},
		ymin=40, ymax=90,
		major tick style={decentgrey}, ytick distance={10}, height=0.2\textheight, enlarge y limits=0, enlarge x limits=0.05]
		\addplot[mark=*,  mark options={solid}, mark size=1.5pt, line width=0.6pt,solid,color=c0] coordinates {
			(1, 81.72)
			(2, 81.86)
			(3, 82.35)
			(4, 83.25)
			(5, 83.55)
			(6, 83.84)
			(7, 84.65)
		};
		\addlegendentry{\setfit student}
		
		\addplot[mark=triangle*, mark options={solid}, mark size=1.5pt, line width=0.6pt,solid,color=c1] coordinates {
			(1, 56.92)
			(2, 62.69)
			(3, 62.27)
			(4, 69.98)
			(5, 74.37)
			(6, 80.55)
			(7, 83.42)
		};
		\addlegendentry{Baseline student}

		\node[anchor=south]() at (4.1, 91) {\small AG News};
		\end{axis}
    \end{tikzpicture}%
    ~%
    %
    \begin{tikzpicture}[]
    		\begin{axis}[
		width=0.37\linewidth, 
		legend style={draw=darkgrey, fill=white!75, text opacity =1, fill opacity=0.8,
		at={(0.97,0.03)},anchor=south east, font=\sffamily\scriptsize},
		legend cell align=left,
		legend columns=1,
		xtick={1,2,3,4,5,6,7},
		xticklabels={8,16,32,64,100,200,1K},
		xlabel={\small Unlabeled Training Set Size ($N$)},
		grid=major, clip=false,
		major grid style={line width=.2pt,draw=decentgrey},
		minor tick style={decentgrey!0},
		x tick label style={/pgf/number format/1000 sep=},
		tick label style={font=\tiny},
		ymin=-5, ymax=70,
		major tick style={decentgrey}, ytick distance={10}, height=0.2\textheight, enlarge y limits=0, enlarge x limits=0.05]
		\addplot[mark=*,  mark options={solid}, mark size=1.5pt, line width=0.6pt,solid,color=c0] coordinates {
			(1, 45.7)
			(2, 46.94)
			(3, 48.97)
			(4, 52.05)
			(5, 54.97)
			(6, 58.81)
			(7, 64.27)
		};
		\addlegendentry{\setfit student}
		
		\addplot[mark=triangle*, mark options={solid}, mark size=1.5pt, line width=0.6pt,solid,color=c1] coordinates {
			(1, 20.59)
			(2, 21.56)
			(3, 21.27)
			(4, 22.86)
			(5, 25.01)
			(6, 35.88)
			(7, 64.48)
		};
		\addlegendentry{Baseline student}

		\node[anchor=south]() at (4.1,71) {Emotion};
		\end{axis}
    \end{tikzpicture}%
    ~%
    %
     \begin{tikzpicture}[] 
     		\begin{axis}[
		width=0.37\linewidth, 
		legend style={draw=darkgrey, fill=white!75, text opacity =1, fill opacity=0.8,
		at={(0.97,0.03)},anchor=south east, font=\sffamily\scriptsize},
		legend cell align=left,
		legend columns=1,
		xtick={1,2,3,4,5,6,7,8,9},
		xticklabels={8,16,32,64,100,200,1K},
		xlabel={\small Unlabeled Training Set Size ($N$)},
		grid=major, clip=false,
		major grid style={line width=.2pt,draw=decentgrey},
		minor tick style={decentgrey!0},
		x tick label style={/pgf/number format/1000 sep=},
		tick label style={font=\tiny},
		ymin=10, ymax=40,
		major tick style={decentgrey}, ytick distance={10}, height=0.2\textheight, enlarge y limits=0, enlarge x limits=0.05]
		\addplot[mark=*,  mark options={solid}, mark size=1.5pt, line width=0.6pt,solid,color=c0] coordinates {
			(1, 30.82)
			(2, 30.72)
			(3, 31.47)
			(4, 32.23)
			(5, 33.12)
			(6, 34.67)
			(7, 38.38)
		};
		\addlegendentry{\setfit student}
		
		\addplot[mark=triangle*, mark options={solid}, mark size=1.5pt, line width=0.6pt,solid,color=c1] coordinates {
			(1, 21.95)
			(2, 23.98)
			(3, 22.93)
			(4, 24.4)
			(5, 24.55)
			(6, 31.45)
			(7, 37.54)
		};
		\addlegendentry{Baseline student}
		\node[anchor=south]() at (4.1,40) {\small SST5};
		\end{axis}
     \end{tikzpicture}
	\caption{Average accuracy as a function of the unlabeled training set size $N$ of the \setfit student and the baseline student on AG News, Emotion and SST5 datasets.}
	\label{fig:distillation_graph}

\vspace{-1em}
\end{figure*}
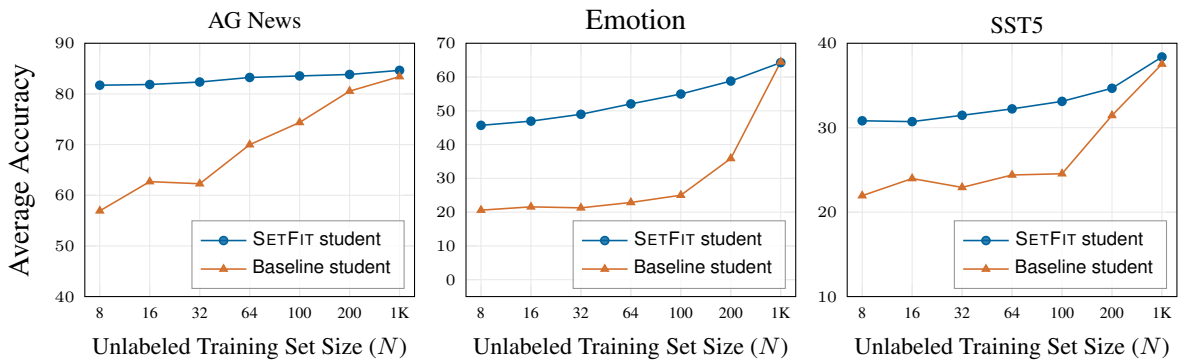

\label{distill_section}
We have shown that \setfit achieves state-of-the-art results in few-shot setups using underlying base models such as paraphrase-mpnet-base-v2 and \robertalarge, containing 110M parameters and 355M parameters respectively; but in real-world deployments, where cost and sustainability are prioritized, the use on even more efficient models is desirable. Previous works have shown model distillation to be effective in reducing computational load while preserving much of the original model’s performance \citep{BA_2014distillation, hinton2015distillation}. In this section we evaluate the performance of \setfit as a student model compared to a standard transformer student model in few-shot distillation setups when the amount of unlabeled training data is limited.

\paragraph{Experimental Setup}
For the distillation tests we use the datasets AGNews, Emotion and SST-5 described in Appendix \ref{appdx:datasets}. 
For the \setfit teacher we chose \setfitbase, which contains 110M parameters, whereas for the \setfit student we chose \setfittiny, which is a much smaller model (15M parameters). For fair comparison, we use as the baseline student MiniLM-L3-H384-uncased\footnote{\url{huggingface.co/nreimers/MiniLM-L3-H384-uncased}}, a standard transformer encoder of the same size as our \setfit student model. For each of the three datasets we train the \setfit teacher model using only 16 labeled samples per class, and the student models are trained using the same 16 labeled samples per class together with various amounts of additional unlabeled data. We follow the same data-split policy and \setfit training parameters' settings described in Section \ref{subsec:main_experimental_setup}.

\paragraph{Method}
The \setfit student is trained using sentence pairs and the level of similarity between each pair as input. The similarity is generated by using the underlying ST of the teacher to produce sentence embeddings for each pair and to calculate the cosine-similarity between them. The underlying ST of the \setfit student is trained to mimic the ST of the teacher output by minimizing the error between the \setfit teacher-produced cosine-similarity and its output. The classification head of the student is then trained using the embeddings produced by the student's ST and the logits produced by the \setfit teacher classification head.
The baseline student is trained to mimic the teacher output by minimizing the error between the logits produced by the \setfit teacher classification head and its output. 

\paragraph{Results}

Figure \ref{fig:distillation_graph} shows a comparison between the \setfit student model and the baseline student model for various amounts of unlabeled training data ($N$). The \setfit student significantly outperforms the baseline student when only small amounts of unlabeled data are available. For example, for $N=8$, the \setfit student outperforms the baseline student by \textbf{24.8}, \textbf{25.1}, and \textbf{8.9} average accuracy on the AGNews, Emotion and SST5 datasets, respectively. As $N$ increases, the performance gains decrease and are on par for $N=1K$.

\subsection{Computational costs}

Comparing the relative computational costs of \setfit versus PET and PEFT methods isn't straightforward since each method typically has different hardware and memory requirements. 

To simplify the comparison, we follow the approach adopted by \citet{t-few} and use FLOPs-per-token estimates to compare \setfit to \tfew. These estimates can be obtained from \citet{scaling-laws}, who show that encoder-only models with $N$ parameters have approximately $2N$ FLOPs-per-token for inference and $6N$ FLOPs-per-token for training. The resulting cost for inference and training is then given by:

\begin{align*}
    {\cal C}_\mathrm{inf} &= 2N\cdot\ell_\mathrm{seq}\,, \\
    {\cal C}_\mathrm{train} &= 6N \cdot \ell_\mathrm{seq} \cdot n_\mathrm{steps} \cdot n_\mathrm{batch}\,,
\end{align*}

where $\ell_\mathrm{seq}$ is the input sequence length, $n_\mathrm{steps}$ is the number of training steps, and $n_\mathrm{batch}$ is the batch size. For encoder-decoder models like \tfew, these estimates are halved, since the model only processes each token with either the encoder or decoder.

For the inference and training estimates shown in Table \ref{table:computation_cost}, we use $\ell_\mathrm{seq} = 38$ and  $\ell_\mathrm{seq} = 54$ as the input sequence length for \setfitbase (\tfew); this is the median number of tokens across all the test datasets in Table~\ref{table:main_results}. We also use $n_\mathrm{steps}=1000$ and $n_\mathrm{batch}=8$ for all training estimates. As shown in the table, the \setfitbase model is approximately an order of magnitude faster at inference and training than \tfew, despite having comparable performance on the test datasets of Table~\ref{table:main_results}. \setfittiny is two orders of magnitude faster than \tfew, with an average score reduction of 3.1 accuracy points. Moreover, the storage cost of the \setfit models (70MB and 420MB respectively) is 163 to 26 times smaller than the \textsc{T0-3B} checkpoint used by \tfewsmall (11.4GB), making these models much better suited for real-world deployment.

These estimates are borne out by comparing the time needed to train each method to convergence on $N=8$ examples. For our datasets, \setfitbase takes approximately 30 seconds to train on a p3.2xlarge AWS instance (16GB GPU memory), at a cost of \$0.025 per split. On the other hand, \tfewsmall requires at least 40GB GPU memory, and training on a p4d.24xlarge AWS instance takes approximately 700 seconds, at a cost of \$0.7 per split.

\begin{table}
\centering
\small
\begin{tabular}{p{2.0cm} p{0.9cm} p{0.9cm} p{0.9cm} p{0.9cm}}
\toprule
\phantom{placeholder} \textbf{Method} & \textbf{Inf. FLOPs} & \textbf{Train FLOPs} & \textbf{Speed-up} & \phantom{placeholder} \textbf{Score} \\
\midrule
\tfewsmall\ & 1.6e11 & 3.9e15 & 1x & 63.4$_{1.9}$ \\
\setfitbase  & 8.3e9 & 2.0e14 & 19x & 62.3$_{4.9}$ \\
\setfittiny{$^\dagger$}  & 1.3e9 & 3.2e13 & 123x & 60.3$_{1.6}$ \\
\bottomrule
\end{tabular}

\caption{\label{table:computation_cost} Relative computational cost and average scores of \setfit and \tfew using $|N|=8$ on the test datasets listed in Table~\ref{table:main_results}. $^\dagger$Trained in the distillation setup as described in Section \ref{distill_section}, using $|N|=8$ for teacher training and the rest of the available training data as unlabeled student training data. For fixed $n_\mathrm{steps}$ and $n_\mathrm{batch}$, the relative speed-up $(N' \cdot \ell'_\mathrm{seq}) / (2N \cdot \ell_\mathrm{seq})$ is the same for inference and training.} 

\end{table}

\section{Conclusion}
This paper introduces \setfit, a new few-shot text classification approach. We show that \setfit has several advantages over comparable approaches such as \tfew, \adapet and \perfect. In particular, \setfit is much faster at inference and training; \setfit requires much smaller base models to be performant, not requiring external compute; and \setfit is additionally not subject to the instability and inconvenience of prompting. We have also demonstrated that \setfit is a robust few-shot text classifier in languages other than English across varying typologies. Finally, \setfit has proven useful in few-shot distillation setups.

\section*{Acknowledgements}
The authors thank Hugging Face Inc and Intel Inc. for providing computing resources and the German Federal Ministry of Education and Research and the Hessian Ministry of Science and the Arts (HMWK) within the projects "The Third Wave of Artificial Intelligence - 3AI",  hessian.AI, and within their joint support of the National Research Center for Applied Cybersecurity ATHENE.

\bibliography{anthology,custom}
\bibliographystyle{acl_natbib}

\appendix

\section{Appendix}
\label{sec:appendix}

\subsection{Datasets}
\label{appdx:datasets}
Table \ref{table:english_dev_data} shows the development and test datasets that are used for setting \setfit's hyperparameters. Following is a description of the datasets used:

\begin{table*}
\centering
\begin{tabular}{lllllll}
\toprule
\textbf{Dataset Name} & \textbf{Type of Task} & \textbf{Cls.}$^*$ & \textbf{Label Dist.{**}} & \textbf{Metric} &\textbf{Split}\\
\midrule

SST5 & Sentiment & 5  & Approx. equal & Accuracy & Test\\
Amazon Counterfactual & Counterfactual & 2 & 10\% counterfactual & MCC & Test \\
CR & Sentiment & 2 & Equal & Accuracy & Test\\
Emotion & Emotion & 6  & Equal & Accuracy & Test\\
Enron Spam & Unwanted Language & 2 & Equal & Accuracy & Test\\
AG News & Topic & 4 & Equal & Accuracy & Test\\
SST2 & Sentiment & 2 & Equal & Accuracy & Dev\\
IMDB & Sentiment & 2 &  Equal & Accuracy & Dev\\
BBC News & Topic & 5 & Equal & Accuracy & Dev\\
Student Question Categories & Topic & 4 &  Approx.Equal & Accuracy & Dev\\
TREC-QC & Topic & 50  & N/A & Accuracy & Dev\\
Toxic Conversations & Unwanted Language & 2 & 8\% Toxic & Avg. Precision & Dev\\
Amazon Polarity & Sentiment & 2 & Equal & Accuracy & Dev\\
\bottomrule
\end{tabular}
\caption{\label{table:english_dev_data}
English datasets used for development and test experiments. $^*$No. of classes per dataset. $^{**}$Distribution of the examples across classes.
}
\end{table*}

\subparagraph{\textbf{SST2}} The Stanford Sentiment Treebank 2 is a collection of single sentence movie reviews with positive-negative sentiment class labels. \cite{socher-etal-2013-recursive}.

\subparagraph{\textbf{IMDB}} The Internet Movie Database dataset is a collection of single sentence movie reviews with positive-negative sentiment class labels. \cite{maas-EtAl:2011:ACL-HLT2011}.

\subparagraph{\textbf{BBC News}} The BBC News dataset is a collection of articles from the news outlet BBC with one of 5 topic classifications: Politics, Sports, Entertainment, Tech, and Business. \cite{bbcNews}.

\subparagraph{\textbf{Enron Spam}} The Enron spam email dataset consists of emails from the internal Enron correspondence channel where emails are classified as spam or not spam. \cite{enron}.

\subparagraph{\textbf{Student Question Categories}\footnote{www.kaggle.com/datasets/mrutyunjaybiswal/iitjee-neet-aims-students-questions-data}} This is a set of questions from university entrance exams in India that are classified into 4 subjects: Math, Biology, Chemistry, Physics.

\subparagraph{\textbf{TREC-QC}} The Text Retrieval Conference Question Answering dataset.

\subparagraph{\textbf{Toxic Conversations}\footnote{https://www.kaggle.com/competitions/jigsaw-unintended-bias-in-toxicity-classification/data}} The Toxic Conversations dataset is set of comments from Civil Comments, a platform for reader comments for independent news outlets. Human raters have given them toxicity attributes. 


\subparagraph{\textbf{Amazon Polarity}\footnote{\url{hf.co/datasets/amazon_polarity}}} The Amazon Polarity dataset consists of customer reviews from \textit{Amazon} taken over 18 years with binary sentiment labels. 
Examples are either positive ("Great Read") or negative ("The Worst!") labelled. \cite{NIPS2015_250cf8b5}.

Following is a description of the test datasets:
\subparagraph{\textbf{Stanford Sentiment Treebank-5 (SST5)}} The SST-5 dataset is the fine-grained version of the Stanford Sentiment Treebank, where each example is given one of five labels: very positive, positive, neutral, negative, very negative.

\subparagraph{\textbf{Amazon Counterfactual}} The Amazon Counterfactual dataset is set of \textit{Amazon} customer reviews with professionally labeled binary labels of counterfactual detection. Counterfactual statements are statements that denote something that did not happen or cannot (e.g. "They are much bigger than I thought they would be."). We used the English subset for our experiments. \cite{amazon_cfd}.

\subparagraph{\textbf{Customer Reviews}}
The Customer Reviews \cite{custo_reviews} dataset is part of the of SentEval \cite{conneau-kiela-2018-senteval} benchmark. It is composed of positive and negative opinions mined from the web and written by customers about a variety of products.

\subparagraph{\textbf{Emotion}\footnote{\url{hf.co/datasets/emotion}}} The Emotion dataset consists of tweets from \textit{Twitter } that display clear emotions (e.g. "i am now nearly finished [with] the week detox and i feel amazing"). Labels are one of six categories: anger, fear, joy, love, sadness, and surprise. \cite{saravia-etal-2018-carer}.

\subparagraph{\textbf{AG News}} AG News is a dataset of news titles from AG news with one of 4 classifications (World, Entertainment, Sports, and Business). \cite{Zhang2015CharacterlevelCN}.

\subsection{\adapet Training Procedure}
\label{appdx:adapet_training_procedure}
By default, \adapet assumes access to a training, development, and test dataset. It trains for $1,000$ batches, runs predictions on the development data every $250$ batches and checkpoints, keeping the model state which performed best on the development dataset. In our case, where we assume few-shot training and no development data, we ran \adapet for $1,000$ batches and disabled the checkpointing, using the model state that resulted after training for $1,000$ batches. For the English data in Table \ref{table:main_results}, we used the pattern "[TEXT1] this is [LBL]", where "[TEXT1]" and "[LBL]" are placeholders for a given piece of text and the corresponding label, respectively. We constructed the verbalizer from the "label" and "label text" columns that are available in all of our datasets. For the multilingual datasets in Table \ref{table:multiling_results}, we used the same pattern, but asked native speakers of each language to translate this pattern into their language. We additionally constructed the verbalizer by mapping labels to a star rating, for example, $0 = 1 \ star$ and $4 = 5 \ stars$, again asking native speakers of each language to translate the verbalizer into their language.

\subsection{Prompts used in \tfew}
\label{appdx:tfew_prompts}
The \textbf{Emotion} dataset is the only one that had existing prompts in P3 (Public Pool of Prompts) \citep{p3}. 
For three other datasets, we had to adapt existing prompts designed for similar datasets on P3, by making minimal required changes to address the differences in data domains or label names:
\begin{itemize}
    \item Prompts for \textbf{Enron Spam}, a spam e-mail detection dataset, were adapted from $\texttt{sms\_spam}$ dataset prompts.
    \item \textbf{CR} prompts were adapted from $\texttt{amazon\_polarity}$.
    \item \textbf{SST5} prompts were adapted from $\texttt{yelp\_review\_full}$. 
\end{itemize}

The \textbf{Amazon Counterfactual} dataset does not have any relevant prompts on P3. Hence, we manually generated prompts ourselves, based on standard practices for prompt creation published in P3. We also added 
two new prompts for \textbf{SST5}, to make it compatible with the label names of \textbf{SST5}.
Following is a list of prompts we created for each dataset:

\paragraph{Amazon Counterfactual Prompts}\mbox{}\\


\footnotesize
Input template:

\texttt{\{\{ text \}\} Is the statement factual?}
\vspace*{0.5cm}

Target template:

\texttt{\{\{ answer\_choices[label] \}\}}
\vspace*{0.5cm}

Answer choices template:

\texttt{Yes ||| No}

\textcolor[RGB]{220,220,220}{\rule{\linewidth}{0.2pt}}


Input template:

\texttt{\{\{ text \}\} Does the statement describe}
\indent\texttt{a fact?}
\vspace*{0.5cm}

Target template:

\texttt{\{\{ answer\_choices[label] \}\}}
\vspace*{0.5cm}

Answer choices template:

\texttt{Yes ||| No}

\textcolor[RGB]{220,220,220}{\rule{\linewidth}{0.2pt}}


Input template:

\texttt{\{\{ text \}\} Is the statement}
\indent\texttt{non-counterfactual or counterfactual?}
\vspace*{0.5cm}

Target template:

\texttt{\{\{ answer\_choices[label] \}\}}
\vspace*{0.5cm}

Answer choices template:

\texttt{non-counterfactual ||| counterfactual}

\textcolor[RGB]{220,220,220}{\rule{\linewidth}{0.2pt}}


Input template:

\texttt{\{\{ text \}\} Is the statement}
\indent\texttt{counterfactual?}
\vspace*{0.5cm}

Target template:

\texttt{\{\{ answer\_choices[label] \}\}}
\vspace*{0.5cm}

Answer choices template:

\texttt{No ||| Yes}

\textcolor[RGB]{220,220,220}{\rule{\linewidth}{0.2pt}}


Input template:

\texttt{\{\{ text \}\} Does the sentence express}
\indent\texttt{an event that did not happen?}
\vspace*{0.5cm}

Target template:

\texttt{\{\{ answer\_choices[label] \}\}}
\vspace*{0.5cm}

Answer choices template:

\texttt{No ||| Yes}

\textcolor[RGB]{220,220,220}{\rule{\linewidth}{0.2pt}}


Input template:

\texttt{\{\{ text \}\} Does this describe an}
\indent\texttt{actual event?}
\vspace*{0.5cm}

Target template:

\texttt{\{\{ answer\_choices[label] \}\}}
\vspace*{0.5cm}

Answer choices template:

\texttt{Yes ||| No}

\textcolor[RGB]{220,220,220}{\rule{\linewidth}{0.2pt}}


Input template:

\texttt{\{\{ text \}\} Does the sentence contain}
\indent\texttt{events that did not or cannot take}
\indent\texttt{place?}
\vspace*{0.5cm}

Target template:

\texttt{\{\{ answer\_choices[label] \}\}}
\vspace*{0.5cm}

Answer choices template:

\texttt{Yes ||| No}

\textcolor[RGB]{220,220,220}{\rule{\linewidth}{0.2pt}}


Input template:

\texttt{Is the label for the following}
\indent\texttt{sentence non-counterfactual or}
\indent\texttt{counterfactual? \{\{ text \}\}}
\vspace*{0.5cm}

Target template:

\texttt{\{\{ answer\_choices[label] \}\}}
\vspace*{0.5cm}

Answer choices template:

\texttt{non-counterfactual ||| counterfactual}

\textcolor[RGB]{220,220,220}{\rule{\linewidth}{0.2pt}}

\paragraph{New prompts for SST5}\mbox{}\\


Input template:

\texttt{How do you feel about the following}
\indent\texttt{sentence? \{\{ text \}\}}
\vspace*{0.5cm}

Target template:

\texttt{\{\{ answer\_choices[label] \}\}}
\vspace*{0.5cm}

Answer choices template:

\texttt{very negative ||| negative ||| neutral}
\indent\texttt{||| positive ||| very positive}

\textcolor[RGB]{220,220,220}{\rule{\linewidth}{0.2pt}}


Input and target templates:

\texttt{\{\{ text \}\} This movie is a very}
\indent\texttt{\{\{answer\_choices[label]\}\} one}
\vspace*{0.5cm}

Answer choices template:

\texttt{terrible ||| bad ||| okay ||| good |||}
\indent\texttt{great}

\textcolor[RGB]{220,220,220}{\rule{\linewidth}{0.2pt}}


\end{document}